\documentclass[conference]{IEEEtran}
\IEEEoverridecommandlockouts
\usepackage{url}

\usepackage{cite}
\usepackage{amsmath,amssymb,amsfonts}
\usepackage{algorithm}
\usepackage{algpseudocode}
\usepackage{csquotes}
\usepackage{graphicx}
\usepackage{textcomp}
\usepackage{xcolor}
\usepackage{placeins}
\usepackage{siunitx}
\def\BibTeX{{\rm B\kern-.05em{\sc i\kern-.025em b}\kern-.08em
    T\kern-.1667em\lower.7ex\hbox{E}\kern-.125emX}}

\usepackage[acronym]{glossaries}

\makeglossaries 

\newacronym{TPPN}{TPPN}{Texture-Penalized Prototype Network}
\newacronym{TPB}{TPB}{Texture-Penalization Branch}

\newacronym{CNN}{CNN}{Convolutional Neural Network}
\newacronym{ViT}{ViT}{Vision Transformer}
\newacronym{DANN}{DANN}{Domain Adversarial Neural Network}
\newacronym{MLP}{MLP}{Multi-Layer Perceptron}

\newacronym{GRL}{GRL}{Gradient Reversal Layer}
\newacronym{SIN}{SIN}{Stylized-ImageNet}

\newacronym{ID}{ID}{In-Distribution}
\newacronym{OOD}{OOD}{Out-of-Distribution}
\newacronym{ODD}{ODD}{Operational Design Domain}

\usepackage[pscoord]{eso-pic}
\newcommand{\arXivNotice}{%
  \AddToShipoutPictureBG*{%
    \put(\LenToUnit{0.5\paperwidth},\LenToUnit{0.96\paperheight}){%
      \makebox[0pt][c]{%
        \footnotesize\sffamily
        \begin{tabular}{c}
          Accepted at the 22nd IEEE International Conference on Advanced Visual and Signal-Based Systems (AVSS 2026). \\
          \copyright~2026 IEEE. Personal use of this material is permitted. Permission from IEEE must be obtained for all other uses.
        \end{tabular}%
      }%
    }%
  }%
}

\begin{document}

\title{Mitigating Shortcut Learning: Texture-Penalized Prototype Networks}

\author{
\IEEEauthorblockN{1\textsuperscript{st} Akshay Anilkumar Girija}
\IEEEauthorblockA{\textit{AI Engineering Department} \\
\textit{Institute for AI Safety and Security}\\
Ulm, Germany \\
akshay.anilkumargirija@dlr.de \\
ORCID: 0000-0002-4384-9739}
\and
\IEEEauthorblockN{2\textsuperscript{nd} Elena Hoemann}
\IEEEauthorblockA{\textit{AI Engineering Department} \\
\textit{Institute for AI Safety and Security}\\
Sankt Augustin, Germany \\
elena.hoemann@dlr.de \\
ORCID: 0000-0001-9315-548X}
\\[0.35cm]
\IEEEauthorblockN{4\textsuperscript{th} Sven Hallerbach}
\IEEEauthorblockA{\textit{AI Engineering Department} \\
\textit{Institute for AI Safety and Security}\\
Sankt Augustin, Germany \\
sven.hallerbach@dlr.de \\
ORCID: 0009-0007-9163-0567}
\and
\IEEEauthorblockN{3\textsuperscript{rd} Frank Köster}
\IEEEauthorblockA{\textit{Institute for AI Safety and Security} \\
Braunschweig, Germany \\
frank.koester@dlr.de }
}

\maketitle
\arXivNotice
\vspace{-0.4cm}
\begin{abstract}
Standard \glspl*{CNN} exhibit severe performance degradation due to a strong inductive texture bias that prioritizes local, high-frequency patterns over global structural shapes. 
This dependency causes confident misclassifications during textural changes or environmental effects. 
To address this flaw, this study introduces the \gls*{TPPN}, a novel architectural framework that shifts this inherent bias without depending on resource-intensive augmented datasets. 
Specifically, a \gls*{TPB} imposes a penalty to suppress the extraction of local texture proxies, forcing the network backbone to discard high-frequency cues and extract purified, shape-biased representations. 
By evaluating similarities within a prototype-based hypersphere derived from the final convolutional features, the approach enforces strict geometric constraints, treating objects as compositions of essential parts to achieve robust classification. 
Evaluations on texture-shape cue-conflict datasets and synthetic noise benchmarks demonstrate the stronger shape bias of this structural disentanglement.
The proposed framework reduces the inherent texture bias of a baseline ResNet-50 from \qty{55.11}{\percent} to \qty{29.73}{\percent}, surpassing the texture-suppression capabilities of an off-the-shelf Vision Transformer (ViT-B/16). 
Furthermore, the approach demonstrates robust generalization under cue-conflict conditions, resisting textural shortcut learning when encountering \gls*{OOD} shapes. 
The model maintains stronger shape accuracy against elevated perturbations. 
On clean validation data, the architecture incurs a minimal drop in accuracy of 0.90 percentage points. 
This provides a structural, efficient solution to \gls*{CNN} texture bias.
\end{abstract}

\section{Introduction}
\label{sec:introduction}

\IEEEPARstart{I}{mage-based} recognition is foundational to many critical systems, including medical imaging and industrial anomaly detection~\cite{anthony2023use, buhler2024domain}.
In automated systems, dependable perception is essential to navigate through dynamic environments~\cite{Bogdoll_2022_CVPR} and safely operate within a specified \gls*{ODD}~\cite{AnilkumarGirija2024}.
However, standard \glspl*{CNN} exhibit structural vulnerabilities in dynamic environments. 
These architectures possess an inductive bias toward texture over true semantic geometry, prioritizing local, high-frequency surface statistics~\cite{geirhos2018imagenet, brendel2019approximating}.

This inductive \enquote{texture bias} causes misclassifications under deceptive textural overlays or environmental shifts, despite underlying object geometry. 
Existing mitigation strategies cause distinct trade-offs. 
Training on stylized datasets (\gls*{SIN}) forces a shape bias by stripping images of natural local textures, which can inadvertently degrade accuracy on standard, unperturbed distributions~\cite{geirhos2018imagenet}.
\glspl*{ViT}~\cite{dosovitskiy2021image, naseer2021intriguing} possess an intrinsic bias toward global shape. 
However,they impose computational overheads preventing deployment in real-time edge applications.

To address these limitations, the Texture-Penalized Prototype Network (\gls*{TPPN}) shifts the inductive bias of standard \glspl*{CNN} from local texture to global shape without relying on specialized external datasets. 
The architecture pairs explicit spatial prototype constraints with a Texture-Penalization Branch (\gls*{TPB}) inspired by \glspl*{DANN}~\cite{ganin2015unsupervised}. 
Instead of passively augmenting the training data, the framework applies on-the-fly perturbations (e.g., speckle noise, blur, color jitter) to create a secondary 4-category texture proxy task. 
Utilizing a \gls*{GRL}, the \gls*{TPB} attempts to classify these augmentations while penalizing the backbone for extracting high-frequency features needed to recognize them. 
This active textural suppression reduces the network's sensitivity to these artificial textures during optimization, compelling deep semantic layers to extract robust, shape-biased representations.

The primary contributions of this work include:

\begin{itemize}
    \item \textbf{Texture-Penalization Branch (\gls*{TPB}):} A \gls*{GRL}-based mechanism that actively suppresses high-frequency textural proxies during feature extraction.
    \item \textbf{\gls*{TPPN} Architecture:} A framework that combines textural suppression with spatial prototypes to achieve attention-level shape bias without needing stylized datasets.
    \item \textbf{Robustness to Texture Deception:} The resulting representations resist cue-conflict conditions and extreme perturbations while preserving clean baseline accuracy.
\end{itemize}


\section{Related Work}
\label{sec:related_work}

Although standard \glspl*{CNN} excel on familiar distributions, they often process images as collections of local patches, exhibiting a strong inductive bias toward high-frequency textures rather than global geometric shapes~\cite{geirhos2018imagenet, brendel2019approximating}. 
This texture bias leads to \enquote{shortcut learning} under deceptive textural overlays.

While training on \gls*{SIN} forces a shape bias~\cite{geirhos2018imagenet}, it is computationally expensive and often degrades standard classification accuracy by disrupting discriminative natural features. 
\gls*{ViT} architectures~\cite{dosovitskiy2021image} and hierarchical variants~\cite{liu2021swin} use self-attention to globally integrate spatial information, thereby establishing a strong inherent shape bias that resists severe textural shifts~\cite{naseer2021intriguing}. 
However, these architectures impose computational overheads that often hinder real-time edge deployment. 

A critical challenge remains: matching the shape bias of Transformers with the efficiency of standard \glspl*{CNN}. 
While \gls*{DANN} introduced the \gls*{GRL} to penalize domain-discriminative features for unsupervised domain adaptation~\cite{ganin2015unsupervised}, the proposed framework repurposes it as a textural penalty head. 
By predicting and penalizing synthetic high-frequency perturbations (e.g., speckle noise, blur), the \gls*{GRL} forces the convolutional backbone to discard superficial textures and rely predominantly on robust geometric shapes. 

Beyond regularizing the backbone, an interpretable prototype network structures the feature space by classifying objects based on spatial similarity to learned prototypical parts~\cite{chen2019looks}. 
Coupling this with the \gls*{GRL}-driven texture penalty encourages the architecture to prioritize shape over texture.

\section{Methodology}
\label{sec:method}

\begin{figure*}[!t]
\centering
\includegraphics[width=\textwidth, trim=0cm 0.2cm 0cm 0.3cm, clip]{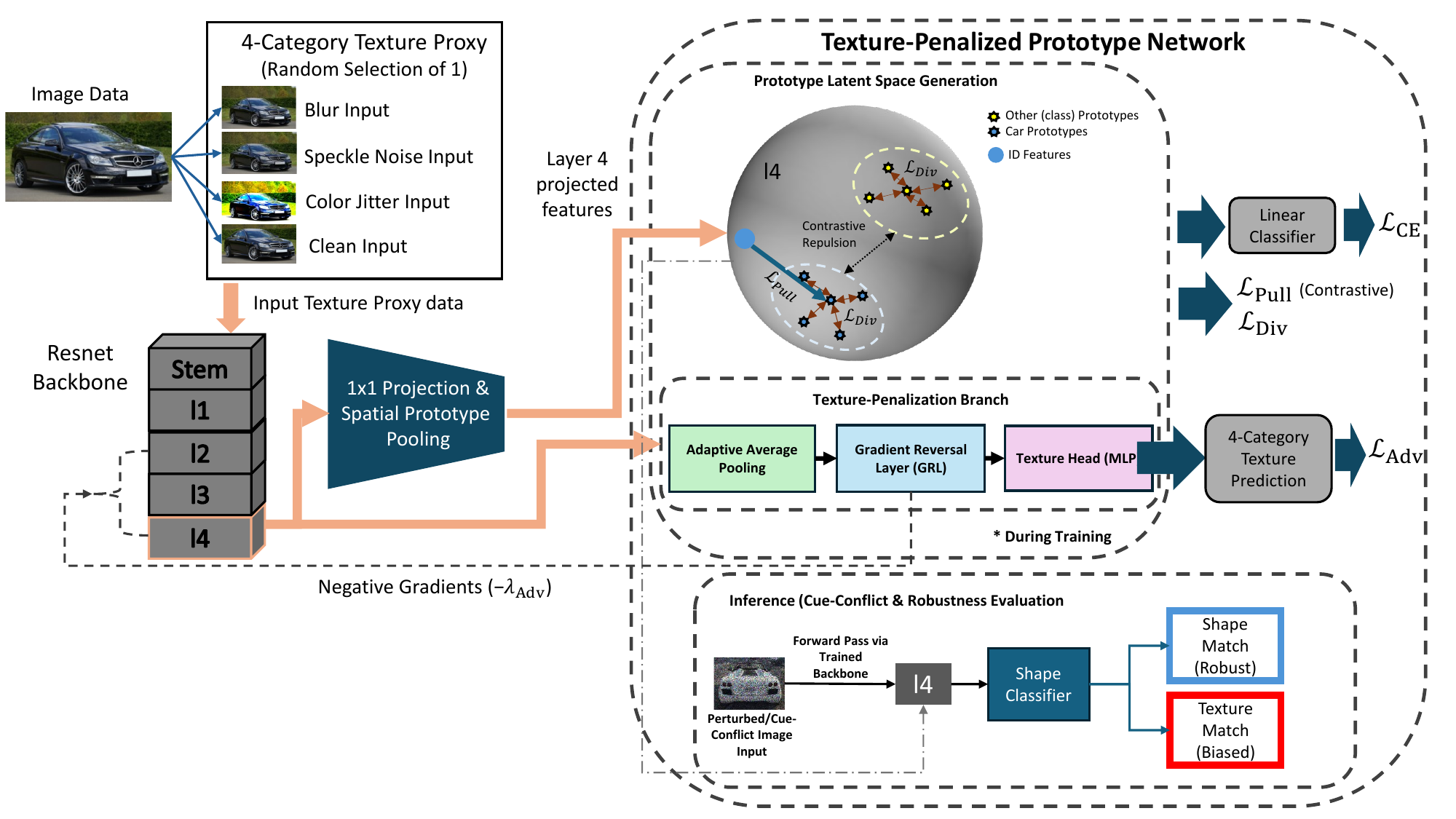}
\caption{Overview of the proposed \gls*{TPPN} framework. 
The network accepts standard image data while applying a singular, random texture proxy augmentation (clean, blur, speckle noise,or color jitter) per image to formulate a 4-category texture training set.
During training (top), the architecture splits deep structural features from Layer 4 (l4) into two distinct pathways. 
In the main classification branch, l4 features are projected onto a shared hypersphere, undergoing spatial prototype matching and pooling to formulate shape-biased predictions guided by Cross-Entropy ($\mathcal{L}_{\mathrm{CE}}$), Pull ($\mathcal{L}_{\mathrm{Pull}}$), and Diversity ($\mathcal{L}_{\mathrm{Div}}$) losses.
In parallel, the \gls*{TPB} isolates raw l4 features to predict the applied texture augmentation. 
A \gls*{GRL} inverts the gradient flow scaled by a penalty factor ($-\lambda_{\mathrm{Adv}}$) back through layers l2 to l4, imposing a penalty on the backbone for recognizing high-frequency texture cues and encouraging the emergence of highly shape-biased representations.
During inference (bottom), the adapted backbone processes cue-conflict or perturbed images. 
The system utilizes the learned shape classifier to quantify structural robustness via the Shape Match metric, contrasting this against residual shortcut learning, evaluated by Texture Match.}
\label{fig:architecture}
\end{figure*}

The \gls*{TPPN} metric-learning architecture decouples geometric shape from surface textures by penalizing high-frequency visual statistics, mapping structural features onto a hypersphere.
As illustrated in Fig. \ref{fig:architecture}, the framework operates through a series of sequential mechanisms, detailed in the following subsections.

\subsection{Deep Semantic Feature Extraction}
Let $\mathbf{X} \in \mathbb{R}^{3 \times H_{\mathrm{in}} \times W_{\mathrm{in}}}$ denote an input image. 
A pre-trained ResNet-50 backbone is utilized for base feature extraction~\cite{he2016deep}. 
To intervene at the level where high-order semantic representations are formed, operations are focused on the deepest convolutional stage, Layer 4 (l4). 
The intermediate feature map $\mathbf{F}_4 \in \mathbb{R}^{C_{\mathrm{in}} \times H \times W}$ is processed by a projection head consisting of a $1\times1$ convolution, batch normalization, and a ReLU activation. 
This yields a dimensionally reduced spatial feature tensor $\mathbf{Z}_4 \in \mathbb{R}^{d \times H \times W}$, where $d$ is the bottleneck projection dimension.

\subsection{Hyperspherical Prototype Matching}
Drawing on part-based prototypical reasoning~\cite{chen2019looks}, a learnable prototype tensor $\mathbf{W} \in \mathbb{R}^{(c \times m) \times d}$ is maintained, where $c$ represents the total target shape classes and $m$ represents the prototypes allocated per class to capture intra-class geometric variance.
Inspired by angular margin formulations that stabilize deep metric boundaries~\cite{wang2017normface, deng2019arcface}, the semantic similarity is decoupled from feature magnitude. Both the spatial features and the prototypes are projected onto a unit $L_2$-hypersphere:
\begin{equation}
\hat{\mathbf{Z}}_4 = \frac{\mathbf{Z}_4}{||\mathbf{Z}_4||_2}, \quad \hat{\mathbf{W}} = \frac{\mathbf{W}}{||\mathbf{W}||_2} .
\label{eq:norm}
\end{equation}
The raw cosine similarity between the spatial feature maps and the prototypes is calculated as the dot product, yielding $S_{\text{raw}} = \hat{\mathbf{Z}}_4 \cdot \hat{\mathbf{W}}$.

\subsection{Dynamic \qty{50}{\percent} Spatial Pooling}
Standard global average pooling aggregates the entire spatial map, diluting the discriminative geometric signal with irrelevant background pixels and localized spatial noise~\cite{he2017mask}. 
To combat this spatial pollution, a dynamic Top-K pooling strategy is introduced, similar to adaptive fractional pooling~\cite{graham2014fractional}, operating on the spatial similarities extracted in Eq. \ref{eq:norm}.
The network applies a descending sort to $S_{\text{raw}}$ across spatial dimensions and averages the top \qty{50}{\percent} of the similarity tokens. The token count, $K$, depends on the input spatial dimensions:
\begin{equation}
K = \max\left(1, \lfloor 0.5 \times H \times W \rfloor\right) .
\label{eq:topk}
\end{equation}
The aggregated similarity matrix $S_{\text{pool, raw}}$ represents the mean of these top $K$ values, isolating object geometry from background interference.

\subsection{Feature Purification via Texture Penalization}
To force the network to discard local textures, a \gls*{TPB} connects to the unprojected l4 feature maps (see Fig. \ref{fig:architecture}, center). 
During training, the pipeline subjects each input image to a stochastic selection of one of four high-frequency synthetic texture proxies: clean (no augmentation), Gaussian blur, speckle noise, or color jitter. 
To process the high-dimensional l4 features, an adaptive average pooling layer reduces the spatial dimensions. 
A \gls*{GRL}~\cite{ganin2015unsupervised} follows the pooling layer. 
After the \gls*{GRL}, a secondary \gls*{MLP} acts as a texture head to predict the applied proxy.

During the forward pass, the \gls*{GRL} acts as an identity function. 
During backpropagation, it multiplies the gradients by a negative scalar ($-\lambda_{\mathrm{Adv}}$). 
The texture head learns to identify the perturbations, while the optimization of the backbone weights $\theta_{\mathrm{backbone}}$ maximizes the texture prediction error:
\begin{equation}
\theta_{\mathrm{backbone}} \leftarrow \theta_{\mathrm{backbone}} + \alpha \lambda_{\mathrm{Adv}} \frac{\partial \mathcal{L}_{\mathrm{Adv}}}{\partial \theta_{\mathrm{backbone}}} ,
\label{eq:grl}
\end{equation}
where $\alpha$ represents the learning rate of the backbone, and $\lambda_{\mathrm{Adv}}$ denotes the gradient reversal penalty scalar. 
By penalizing the backbone for the extraction of high-frequency signals necessary to identify noise or blur, the bottleneck forces the latent space to discard natural textures and rely on global shapes~\cite{geirhos2018imagenet}.

\subsection{Temperature Scaling and Reasoning Head}
To control the sharpness of the similarity distribution during contrastive learning, a critical dynamic for bounded hyperspherical embeddings~\cite{wu2018unsupervised}, $S_{\text{pool, raw}}$ is scaled by a learnable temperature parameter $\tau$. 
A softplus activation with a configurable floor ($\tau_{\text{floor}}$) ensures numerical stability:
\begin{equation}
S_{\text{scaled}} = \frac{S_{\text{pool, raw}}}{\text{Softplus}(\tau) + \tau_{\text{floor}}} ,
\label{eq:temp}
\end{equation}
where $\tau$ receives an initialization of $1.0$, and $\tau_{\text{floor}}$ is set to $0.02$ to prevent division by zero.

To map the similarities into final class predictions without distorting the underlying cosine geometry, a bias-free linear layer is applied to the raw similarities. 
Drawing on the principles of imprinted weights~\cite{qi2018low}, connections between a class logit and its assigned prototypes receive an initialization of $1.0$, while non-target connections receive an initialization of $-0.1$.
This yields the final shape prediction vector, $\mathbf{z}_{\mathrm{shape}}$.

\subsection{The Composite Objective}
Optimization of the framework utilizes a composite loss function, balancing discriminative shape learning against texture suppression:

\begin{equation}
\begin{split}
\mathcal{L}_{\mathrm{Total}} &= \lambda_{\mathrm{CE}}\mathcal{L}_{\mathrm{CE}} + \lambda_{\mathrm{Pull}}\mathcal{L}_{\mathrm{Pull}} \\
&\quad + \lambda_{\mathrm{Div}}\mathcal{L}_{\mathrm{Div}} + \lambda_{\mathrm{Adv}}\mathcal{L}_{\mathrm{Adv}} .
\end{split}
\label{eq:loss}
\end{equation}
\noindent \textbf{Cross-Entropy ($\mathcal{L}_{\mathrm{CE}}$):} The standard classification loss applied to $\mathbf{z}_{shape}$.\\
\textbf{Contrastive Pull ($\mathcal{L}_{\mathrm{Pull}}$):} Adapted from Supervised Contrastive Learning~\cite{khosla2020supervised}, an Information Noise-Contrastive Estimation (InfoNCE) based push-pull dynamic~\cite{oord2018representation} is formulated on the scaled similarities.
Using Dynamic Dropout with a margin $\delta$, target prototypes that fall below a similarity threshold are masked to force the utilization of all $m$ prototypes. 
The loss is calculated against the best valid positive prototype $s^+_{\mathrm{best}}$:
\begin{equation}
\mathcal{L}_{\mathrm{Pull}} = -\log \frac{\exp(s^+_{\mathrm{best}})}{\exp(s^+_{\mathrm{best}}) + \sum \exp(s^-)} .
\label{eq:pull}
\end{equation}
\textbf{Diversity ($\mathcal{L}_{\mathrm{Div}}$):} To prevent feature collapse and ensure broad spatial coverage~\cite{bansal2018can}, this loss enforces orthogonality among the $m$ prototypes within the same class:
\begin{equation}
\mathcal{L}_{\mathrm{Div}} = \sum_{k=1}^c \sum_{i \neq j} (\hat{\mathbf{W}}_{k, i} \cdot \hat{\mathbf{W}}_{k, j})^2 .
\label{eq:div}
\end{equation}
\textbf{Texture-Penalization Loss ($\mathcal{L}_{\mathrm{Adv}}$):} The cross-entropy loss calculated over the 4-class texture predictions, optimized via the \gls*{GRL} mechanism defined in Eq. \ref{eq:grl}.

\section{Experiments}
\label{sec:experiments}

An empirical testing paradigm is established to evaluate the proposed \gls*{TPPN}. 
\subsection{Datasets and Evaluation Metrics}
\textbf{Datasets.} The framework is evaluated across three distinct configurations to isolate shape and texture dependencies. 
\textit{ImageNet dataset ~\cite{deng2009imagenet} (Training):} Models are trained on a balanced 10-class ImageNet subset containing 20,000 total images (2,000 images per class). 
These 10 categories define the \gls*{ID} classes: airplane, bear, bicycle, bird, boat, bottle, car, cat, chair, and clock. 
To generate the 4-category texture proxy dataset, images are augmented during training with four states: clean, Gaussian blur, speckle noise, or color jitter. 
\textit{ImageNet Validation dataset (Testing):} A clean set of 1,600 unperturbed \gls*{ID} images (100--200 per class) is utilized to verify that the structural intervention preserves baseline performance on natural distributions. 
\textit{Cue-Conflict Dataset (Testing)~\cite{geirhos2018imagenet}:} This dataset contains style-transferred images with conflicting shape and texture features. 
These images present a contradiction between semantic geometry and surface patterns (e.g., the shape of a cat rendered with the texture of an elephant). 
The geometric shape defines the correct target class. 
To test vulnerability to texture deception and generalization, six unobserved categories define the \gls*{OOD} classes: dog, elephant, keyboard, knife, oven, and truck. 
From the initial 1,280 images, 230 images were removed because they share identical shape and texture categories. 
The filtered dataset contains 1,050 images divided into three subsets: 450 images with \gls*{ID} shapes and \gls*{ID} textures, 300 images with \gls*{ID} shapes and \gls*{OOD} textures, and 300 images with \gls*{OOD} shapes and \gls*{ID} textures.
To assess structural robustness, the evaluation applies escalating levels of speckle noise ($\eta \in \{\qty{0}{\percent}, \qty{25}{\percent}, \qty{50}{\percent}, \qty{75}{\percent}\}$) to the filtered cue-conflict images.

\textbf{Metrics.} Classification performance is assessed via Top-1 and Top-3 accuracies on the clean validation dataset. 
Inductive bias is quantified on the cue-conflict dataset using three metrics. 
\textit{Shape Accuracy (\gls*{ID} Texture)} the percentage of predictions matching the underlying \gls*{ID} shape when paired with a conflicting \gls*{ID} texture.
\textit{Texture Bias} measures the percentage of classifications aligning with the conflicting ID texture instead of the true object shape.
\textit{Shape Accuracy (\gls*{OOD} Texture)} evaluates shape recognition when the conflicting texture originates from \gls*{OOD} classes.
\textit{Texture Deception Vulnerability (Fooled by Texture)} measures the rate at which the model is fooled by an \gls*{ID} texture overlaid on an \gls*{OOD} shape.

\subsection{Baselines and Ablation Configurations}
The primary baseline is a standard ResNet-50 trained on the \gls*{ID} classes. 
A second baseline, ResNet-50 with \gls*{TPB}, evaluates the texture penalization branch in isolation to demonstrate the necessity of prototypical reasoning. 
Two ImageNet-1K pre-trained models establish external benchmarks: VGG-16 \cite{simonyan2014very} (texture-dependent) and \gls*{ViT}-B/16 \cite{dosovitskiy2021image, naseer2021intriguing} (shape-biased). 
Comparing these generalized, large-scale models against specialized 10-class architectures could introduce a evaluation mismatch. 
To enable zero-shot evaluation without retraining, model outputs require adaptation. 
The 1,000 fine-grained ImageNet classes are mapped into the 10 broad \gls*{ID} categories, where the highest confidence score within each group serves as the final category prediction. 
To contrast these external benchmarks, an ablation model (\gls*{TPPN} Warmup checkpoint) evaluates the network with prototypes trained on semantics without \gls*{TPB} influence.

\subsection{Implementation Details}
Structural features for the proposed \gls*{TPPN} are extracted from Layer 4 of the ResNet-50 backbone. 
These are compressed via a $1 \times 1$ projection to a dimension of $d=64$, with $m=5$ learnable prototypes allocated per class. 
Classification loss weights are set to $\lambda_{\mathrm{CE}}=1.0$, $\lambda_{\mathrm{Pull}}=0.5$, and $\lambda_{\mathrm{Div}}=0.75$. 
Dynamic dropout employs a vicinity margin $\delta=0.15$ with a probability $p=0.65$. 
The penalization weight ($\lambda_{\mathrm{Adv}}$) is scaled via a sigmoid schedule ($0.0 \to 1.0$) to prevent early backbone destabilization. 

Training utilizes Stochastic Gradient Descent (batch size 48, momentum 0.9, weight decay $10^{-4}$) via a two-phase protocol. 
A 10-epoch \textbf{Warmup Phase} freezes the backbone, optimizing the projection, prototype, and reasoning heads at a learning rate of $10^{-3}$. 
This initial freezing establishes stable prototypical decision boundaries prior to the introduction of penalization gradients. 
A 30-epoch \textbf{Texture Penalization Phase} unfreezes the backbone for joint optimization with the texture head via the \gls*{TPB}. 
Differential learning rates (backbone $10^{-4}$, heads $10^{-3}$) are applied and managed by a ReduceLROnPlateau scheduler with a 5-epoch patience. 
Early stopping halts training upon the stagnation of a composite metric representing the difference between validation accuracy and texture accuracy.

\section{Results and Discussion}
\label{sec:results}

For reproducibility, \gls*{TPPN} performance metrics on the cue-conflict dataset are reported as the mean and standard deviation across five runs with distinct random seeds. Unless otherwise specified, baselines reflect single-run evaluations, and pre-trained models are evaluated zero-shot.

\subsection{Performance on ImageNet Validation Dataset}

Table~\ref{tab:clean_validation} presents the classification results on the clean ImageNet validation subset. 
The baseline ResNet-50 achieves a Top-1 accuracy of 99.06\,\% (Top-3: 99.81\,\%). 
The proposed \gls*{TPPN} achieves a Top-1 accuracy of (98.16 $\pm$ 0.20)\,\% (Top-3: (99.62 $\pm$ 0.10)\,\%).
This 0.90\,\% difference in standard accuracy demonstrates that the \gls*{TPB} forces the network to extract shape features without destroying baseline classification capabilities on natural image distributions.

\begin{table}[!htbp]
\centering
\caption{Classification Accuracy on Clean ImageNet Validation Dataset}
\label{tab:clean_validation}
\resizebox{\columnwidth}{!}{%
\begin{tabular}{lcc}
\hline
\textbf{Model Architecture} & \textbf{Top-1 Accuracy (\%)} & \textbf{Top-3 Accuracy (\%)} \\
\hline
ResNet-50 (Baseline) & 99.06 & 99.81 \\ 
TPPN (Final Model) & 98.16 $\pm$ 0.20 & 99.62 $\pm$ 0.10 \\
\hline
\end{tabular}%
}
\end{table}

\subsection{Texture-Shape Disentanglement on Cue-Conflict Data}

Initial evaluations on the cue-conflict dataset utilize clean conditions (\qty{0.0}{\percent} speckle noise) to measure inherent bias without any synthetic high-frequency perturbation.

\begin{table}[!htbp]
\centering
\caption{Evaluation on Clean Cue-Conflict Dataset (0\% Speckle Noise)}
\label{tab:results_cue_conflict}
\setlength{\tabcolsep}{3pt} 
\footnotesize 
\begin{tabular}{@{}lcccc@{}} 
\hline
\textbf{Model} & \textbf{Shape Acc} & \textbf{Texture} & \textbf{Shape Acc} & \textbf{Fooled} \\ 
\textbf{Architecture} & \textbf{(ID Tex) \%} & \textbf{Bias \%} & \textbf{(OOD Tex) \%} & \textbf{by Tex \%} \\ \hline
VGG-16 (Pretrained) & 8.22 & 76.00 & 21.00 & 78.00 \\ 
ResNet-50 (Baseline) & 19.11 & 55.11 & 22.00 & 70.33 \\ 
ResNet-50 (w/ \gls*{TPB}) & 19.11 & 48.22 & 22.33 & 56.67 \\ 
ViT-B/16 (Pretrained) & \textbf{33.78} & 39.78 & \textbf{31.67} & 54.67 \\ \hline
TPPN (w/o \gls*{TPB}) & \begin{tabular}[c]{@{}c@{}}16.22 \\[-0.7ex] {\scriptsize $\pm$ 0.61}\end{tabular} & \begin{tabular}[c]{@{}c@{}}62.76 \\[-0.7ex] {\scriptsize $\pm$ 1.46}\end{tabular} & \begin{tabular}[c]{@{}c@{}}23.73 \\[-0.7ex] {\scriptsize $\pm$ 0.57}\end{tabular} & \begin{tabular}[c]{@{}c@{}}73.40 \\[-0.7ex] {\scriptsize $\pm$ 1.95}\end{tabular} \\ 
\textbf{TPPN (Final)} & \begin{tabular}[c]{@{}c@{}}29.82 \\[-0.7ex] {\scriptsize $\pm$ 0.81}\end{tabular} & \begin{tabular}[c]{@{}c@{}}\textbf{29.73} \\[-0.7ex] \textbf{{\scriptsize $\pm$ 1.55}}\end{tabular} & \begin{tabular}[c]{@{}c@{}}30.13 \\[-0.7ex] {\scriptsize $\pm$ 0.62}\end{tabular} & \begin{tabular}[c]{@{}c@{}}\textbf{40.47} \\[-0.7ex] \textbf{{\scriptsize $\pm$ 2.40}}\end{tabular} \\ \hline
\end{tabular}
\end{table}

As shown in Table~\ref{tab:results_cue_conflict}, the VGG-16 and ResNet-50 baseline demonstrate a dependency on local texture, exhibiting texture biases of \qty{76.00}{\percent} and \qty{55.11}{\percent}. 
Under deceptive conditions where the network observes an \gls*{OOD} shape overlaid with a familiar \gls*{ID} texture, the ResNet-50 baseline is \enquote{fooled} by the texture \qty{70.33}{\percent} of the time. 
The pre-trained \gls*{ViT}-B/16 achieves the highest \gls*{ID} shape accuracy at \qty{33.78}{\percent}, reflecting its architectural shape bias.

Adding the \gls*{TPB} to the baseline (ResNet-50 with \gls*{TPB}) reduces the deception failure rate from \qty{70.33}{\percent} to \qty{56.67}{\percent}. 

The \gls*{TPPN} ablation without the \gls*{TPB} yields a \qty{73.40 \pm 1.95}{\percent} deception failure rate, suggesting that prototype layers alone are insufficient to fully mitigate texture dependency without the \gls*{TPB} branch.
An analysis of these isolated components establishes a structural synergy. The \gls*{TPB} alone is limited in maximizing shape accuracy (\qty{19.11}{\percent}), and prototypes alone struggle to adequately suppress texture bias ($\sim\qty{73.40}{\percent}$ failure rate). Therefore, the architecture benefits significantly from the combination of both mechanisms to achieve robust spatial reasoning and texture suppression.

By combining the prototype architecture with the \gls*{TPB}, a realignment of the network's inductive bias is observed. 

The proposed \gls*{TPPN} (Final Model) reduces the texture bias to \qty{29.73 \pm 1.55}{\percent} and improves \gls*{ID} shape accuracy from \qty{19.11}{\percent} (ResNet-50 baseline) to \qty{29.82 \pm 0.81}{\percent}.

The deception failure rate drops to \qty{40.47 \pm 2.40}{\percent}, indicating that the network resists familiar textures when they do not align with coherent global geometry.

Furthermore, when evaluating known \gls{ID} shapes rendered with unfamiliar \gls*{OOD} textures, the \gls{TPPN} preserves more of its geometric reasoning, achieving a shape accuracy of \qty{30.13 \pm 0.62}{\percent} compared to the ResNet-50 baseline's \qty{22.00}{\percent}.

\begin{figure}[!htbp]
    \centering
    \includegraphics[width=0.9\columnwidth]{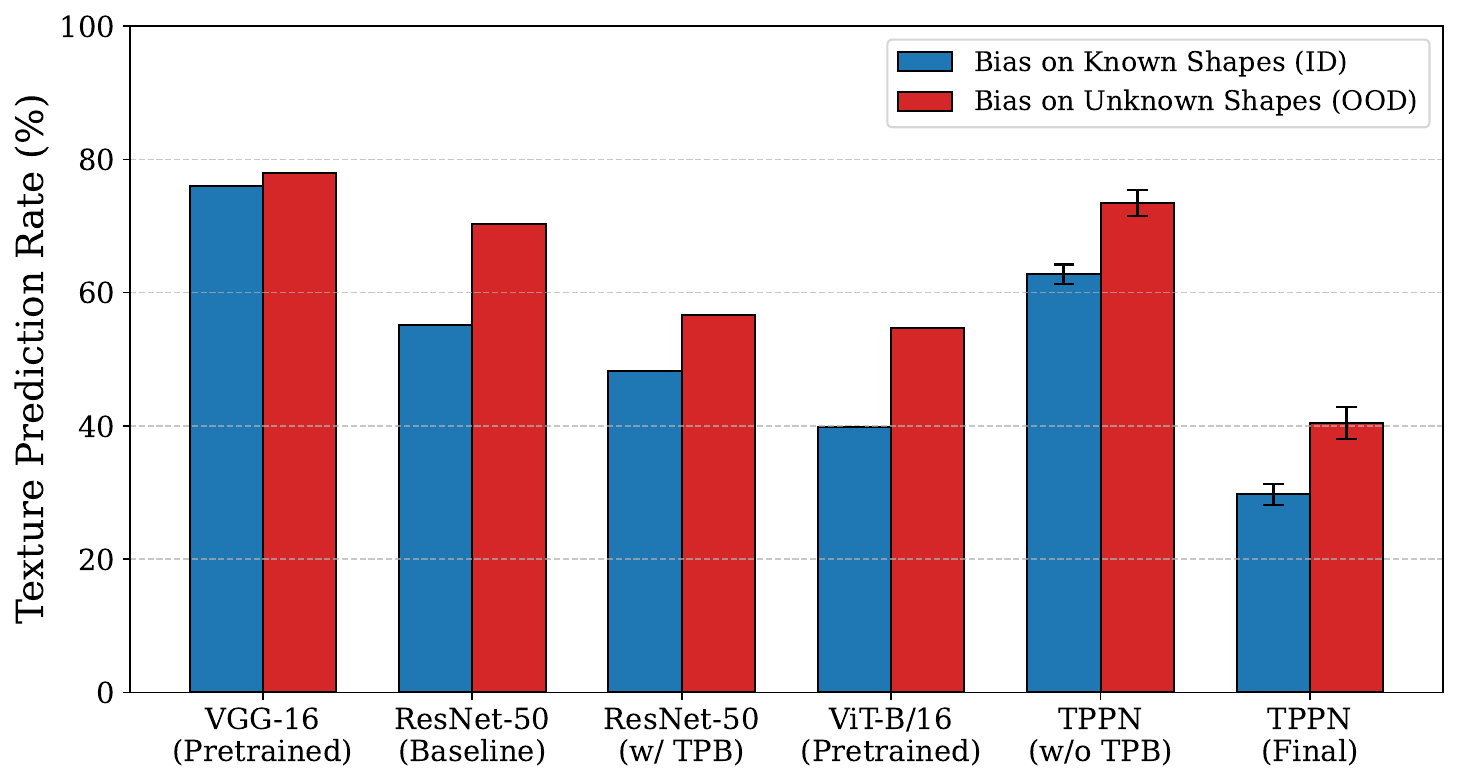}
    \caption{Texture prediction rates across \gls*{ID} and \gls*{OOD} shapes. When evaluating cue-conflict images with unfamiliar \gls*{OOD} shapes, baseline models increase their dependence on familiar textures. The proposed \gls*{TPPN} exhibits reduced fluctuations between clean and deceptive conditions.}
    \label{fig:texture_vuln}
\end{figure}

Figure~\ref{fig:texture_vuln} illustrates the stability of each model when evaluating \gls*{OOD} shapes. 
When the ResNet-50 baseline encounters a known texture overlaid on an unknown \gls*{OOD} shape, its dependency on texture increases from \qty{55.11}{\percent} to \qty{70.33}{\percent}. 
The \gls*{ViT}-B/16 is susceptible to this texture pull, experiencing a 14.89 percentage points increase in texture bias. 
The proposed \gls*{TPPN} restricts this increase to 10.74 percentage points (from \qty{29.73}{\percent} to \qty{40.47}{\percent}), demonstrating a consistent reasoning process across different shape distributions.

The ablation model isolates the source of this consistency. 
Without the penalization branch, texture bias reaches \qty{62.76 \pm 1.46}{\percent} on known shapes and \qty{73.40 \pm 1.95}{\percent} on unfamiliar geometries.
The penalization mechanism substantially reduces these dependencies, prioritizing geometry over localized patterns. 
While the \gls*{ViT}-B/16 achieves the highest shape accuracy on clean \gls*{ID} images (\qty{33.78}{\percent}), the \gls*{TPPN} on a standard ResNet backbone achieves a lower baseline texture bias of \qty{29.73 \pm 1.55}{\percent} compared to \qty{39.78}{\percent}, and is fooled by deceptive textures at a lower rate of \qty{40.47 \pm 2.40}{\percent} compared to \qty{54.67}{\percent}.

This indicates that \glspl*{CNN}, when constrained by high-frequency penalization, can exhibit texture-resilience characteristics comparable to, or exceeding, those of attention-based architectures.

\subsection{Shape Resilience Under Texture Perturbation}

To evaluate shape accuracy (\gls*{ID} texture - \gls*{ID} shape) under texture perturbation, the cue-conflict dataset is subjected to Gaussian speckle noise ($\eta \in \{\qty{0}{\percent}, \qty{25}{\percent}, \qty{50}{\percent}, \qty{75}{\percent}\}$). This forces the model to maintain geometric reasoning, as the noise degrades high-frequency textural patterns while leaving spatial geometries intact.

\begin{figure}[!htbp]
    \centering
    \includegraphics[width=0.9\columnwidth]{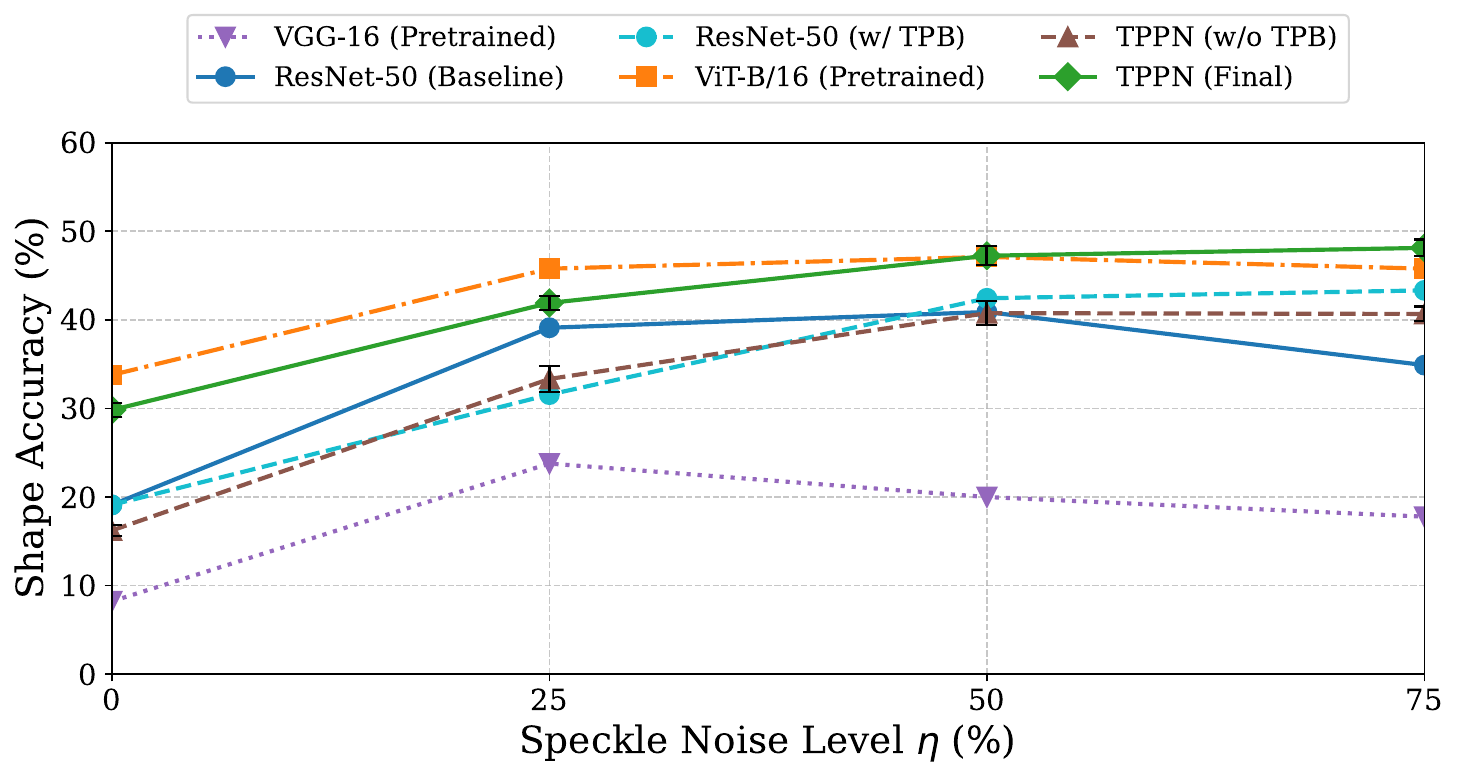}
    \caption{Shape accuracy trajectories under increasing speckle noise. As textures are perturbed, the \gls*{TPPN} relies on global geometry, attaining higher accuracy than baseline architectures at extreme noise levels.}
    \label{fig:noise_line}
\end{figure}

As depicted in Figure~\ref{fig:noise_line}, at moderate perturbation ($\eta=\qty{50}{\percent}$), the ResNet-50 baseline yields \qty{40.89}{\percent} shape accuracy.
The ResNet-50 baseline with the \gls*{TPB} improves this baseline to \qty{42.44}{\percent}.

The proposed \gls*{TPPN} processes the remaining spatial geometries to reach a shape accuracy of \qty{47.24 \pm 1.06}{\percent}.

Under extreme perturbation ($\eta=\qty{75}{\percent}$), the \gls*{TPPN} architecture reaches \qty{48.13 \pm 0.94}{\percent} shape accuracy, surpassing both the ResNet-50 baseline (\qty{34.89}{\percent}) and the \gls*{ViT}-B/16 (\qty{45.78}{\percent}).
This trajectory provides empirical evidence that penalizing local perturbations via the \gls*{TPB} during training mitigates texture bias under clean conditions and encourages a spatial representation that is more resilient to perturbation.

\section{Conclusion and Future Work}
\label{sec:conclusion}

This paper introduces the Texture-Penalized Prototype Network, a structural approach to overcome the inductive texture bias in \glspl*{CNN}. 
By utilizing a Texture-Penalization Branch to penalize the recognition of synthetic image perturbations, the network is forced to discard high-frequency textural shortcuts and learn robust geometric shapes. 
This penalization bottleneck reduces dependency on texture by half, without requiring expensive stylized datasets. 
Compared to off-the-shelf \gls*{ViT} baselines, the resulting architecture exhibits stronger texture suppression and demonstrates resilience to severe image perturbations within the controlled evaluation setting.

Future work will explore deeper \gls*{MLP} configurations for the texture head to capture complex noise patterns. 
Building upon these architectural refinements, scaling the framework to all ImageNet-1K \cite{deng2009imagenet} classes will be essential.
While the restricted class subset was necessary to strictly isolate shape-texture conflicts, evaluating the full dataset alongside finer-grained hyperparameter ablations will be vital to validate the model's broad applicability.

\section*{Acknowledgment}
The authors utilized LLMs strictly for editing and formatting. All scientific content is entirely original.
\FloatBarrier %


\bibliographystyle{IEEEtran.bst}
\bibliography{bibliography.bib}

\end{document}